\documentclass[12pt, reqno]{amsart}
\usepackage[table, dvipsnames]{xcolor}
\usepackage{times}
\usepackage{amssymb, amsmath}
\usepackage{amsthm}
\usepackage{tikz}
\usepackage{bm}
\usetikzlibrary{matrix,arrows}
\usepackage{enumitem}
\usepackage{booktabs}
\usepackage[section]{placeins}
\usepackage{subcaption}
\usepackage{multirow}
\usepackage{hyperref}
\usepackage{float}

\theoremstyle{definition}

\theoremstyle{remark}

\numberwithin{equation}{section}

\begin{document}
\setcounter{page}{1}

\centerline{}

\centerline{}

\title[Lightweight Fish Classification Model]{Lightweight Fish Classification Model for Sustainable Marine Management: Indonesian Case}

\author[F. Kurniawan]{Febrian Kurniawan$^{1}$}

\address{$^{2}$ School of Computing, Telkom University, Republic of Indonesia}
\email{\textcolor[rgb]{0.00,0.00,0.84}{febrian@ieee.org}}

\author[G. B. Satrya]{Gandeva Bayu Satrya$^{2}$}

\address{$^{2}$ Department of Computer Engineering, Canadian University Dubai, Dubai, UAE}
\email{\textcolor[rgb]{0.00,0.00,0.84}{gandevabs@cud.ac.ae}}

\author[F. Kamalov]{Firuz Kamalov$^{3}$}

\address{$^{3}$ Department of Electrical Engineering, Canadian University Dubai, Dubai, UAE.}
\email{\textcolor[rgb]{0.00,0.00,0.84}{firuz@cud.ac.ae}}

\keywords{fish classification; fish dataset; convolutional neural network; data augmentation; machine learning; MobileNets; resource-constrained devices;   sustainable fisheries management; Indonesia}


\begin{abstract}
The enormous demand for seafood products has led to exploitation of marine resources and near-extinction of some species. In particular, overfishing is one the main issues in sustainable marine development. In alignment with the protection of marine resources and sustainable fishing, this study proposes to advance fish classification techniques that support identifying protected fish species using state-of-the-art machine learning. We use a custom  modification of the MobileNet model to design a lightweight classifier called M-MobileNet that is capable of running on limited hardware. 
As part of the study, we compiled a labeled dataset of 37,462 images of fish found in the waters of the Indonesian archipelago. The proposed model is trained on the dataset to classify images of the captured fish into their species and give recommendations on whether they are consumable or not. 
Our modified MobileNet model uses only 50\% of the top layer parameters with about 42\% GTX 860M utility and achieves up to 97\% accuracy in fish classification and determining its consumability. Given the limited computing capacity available on many fishing vessels, the proposed model provides a practical solution to on-site fish classification. In addition, synchronized implementation of the proposed model on multiple vessels can supply valuable information about the movement and location of different species of fish.
\end{abstract} \maketitle

\section{Introduction}
Fish and seafood are among the most highly marketed foods in the world. According to WWF's report \cite{high2014,wwf2020}, over 740 million people (10\%) are reliant on catching, measuring, producing, and selling fish and seafood, and the statistics are continuously growing. People in developing maritime countries are largely dependent on fish as their primary livelihood, distributing the largest volume of fish catch and production worldwide and contributing 97\% of the world’s fishing workforce \cite{kelleher2012}. This also applies to the overwhelming majority of small-scale fishermen for whom fishing makes up the basis of their earnings as well as an essential part of their daily nourishment.

The oceans are home to more than 20,000 species of fish \cite{oceana2020} with some of them consumable and others not. Continuous overfishing of the sea resources not only endangers many fish species, but also the balance of the entire ecosystem. Utilization of an intelligent classification system for fishing will help the fishermen separate endangered fish species from their catches to avoid illegal activity and protect the species. The existing state-of-the-art fish classification models considered limited species of fish \cite{cottrell2020} and did not assess  fish consumability status despite its importance \cite{ruiz2020}. This consumability could be a strong indicator in distinguishing various protected and dangerous edible fish species from their commercial and consumable counterparts.  
	
This work considers fish classification over a remote fishing environment in Indonesia and develops an edge intelligence (EI) strategy to overcome unstable network connections in the sea’s remote areas. The framework consists of state-of-the-art lightweight machine learning (ML) model based on MobileNets and low communication-overhead ML libraries for resource-constrained edge devices (e.g., smartphones). Compared to the existing approaches, we propose a practical solution that can accommodate massive-scale and heterogeneous Internet of Things (IoT) deployments. 
	
Unlike simple batch-based learning approaches, we aim for the development of portable ML libraries for local computation with restricted dependency on remote computing libraries. The proposed method is designed to be implemented on a low and small resource compact machine in remote areas. In particular, we utilize GTX 860m graphics card manufactured in 2014 with 28 nm chip size (low specification mobile chip with low-resource computing for remote area implementation). This resource-efficient feature of the strategy is contributed by possible customization according to the characteristics of the target of the EI application.
	
Although evaluating species could be time-consuming and largely laborious, it is a mandatory procedure for both industrial and research fishing boats. Aboard research fishing boats, the fish species are often evaluated manually. For example, the length of the fish is estimated manually with one person measuring the length of the fish using a measuring board, while another person manually writes the data and records it in a personal computer \cite{strachan1994}. Automatic fish length measurement in the laboratory by using computer vision methods has been considered in  \cite{arnarson1988,strachan1993}, which showed results of less than 1 cm of errors. In a related study the authors  used the combination of a Bayes maximum likelihood classifier, a Learning Vector Quantification (LVQ) classifier, and a One-Class-One-Network (OCON) neural network classifier to develop an intelligent system for counting fish by species \cite{Cadieux2000}.

There has been a growing interest in fish classification in the recent literature  \cite{salman2016,taheri2020,perez2017,sheaves2020}. 
In particular, several approaches to fish classification have used deep learning models \cite{rathi2017,chen2017,alvarez2020,rauf2019,xu2018,garcia2020,iqbal2019,cui2020,zhao2020,tao2019}.  
Different classification approaches were introduced and could be generally categorized into MobileNets \cite{howard2017}, MobileNetv2 \cite{sandler2018}, VGG16 \cite{simonyan2014}, Resnet50 \cite{he2016}, Effnet \cite{freeman2018}, Capsnet \cite{sabour2017}, Sufflenet \cite{zhang2018}, Mnasnet \cite{tan2019}, Xception \cite{chollet2017}. MobileNets are one of the promising candidates for the next deep learning method in object detection and classification for  large datasets. MobileNets \cite{howard2017,sandler2018} rely on a streamlined architecture that applies depth-wise separable convolutions to make lightweight deep neural networks. MobileNets has been implemented in many applications such as traffic density \cite{biswas2019}, redundancy reduction  \cite{su2018}, skin classification \cite{lim2019}, FPGA \cite{shen2019}, vehicle counting \cite{heredia2019}, multi-fruit detection \cite{basri2018}, fish species classification \cite{hung2020}, object detection on non-GPU \cite{sanjay2019}, and reduced MobileNetv2 \cite{ayi2020}.

In this research, we propose an enhanced MobileNet model called Modified-MobileNet (M-MobileNets) that focuses on reducing the top layer of the CNN and improving the accuracy of deployment in low-specification devices (i.e., GeForce GTX 860M). Due to the distinct population of Indonesian fish species, we build our own  fish dataset that includes the images and species labels of 37,462 specimens of fish together with the consumability index. We extract the fish species status based on the poisonous, traumatogenic, and venomous characteristics shared by the fisherfolk. The dataset contains a total of 667 different species of fish. The dataset is used to train the proposed M-MobileNet model. The results show that M-MobileNet outperforms the existing benchmark methods.

\begin{figure}[!t]
		\centering
		\includegraphics [width=0.25\textwidth]{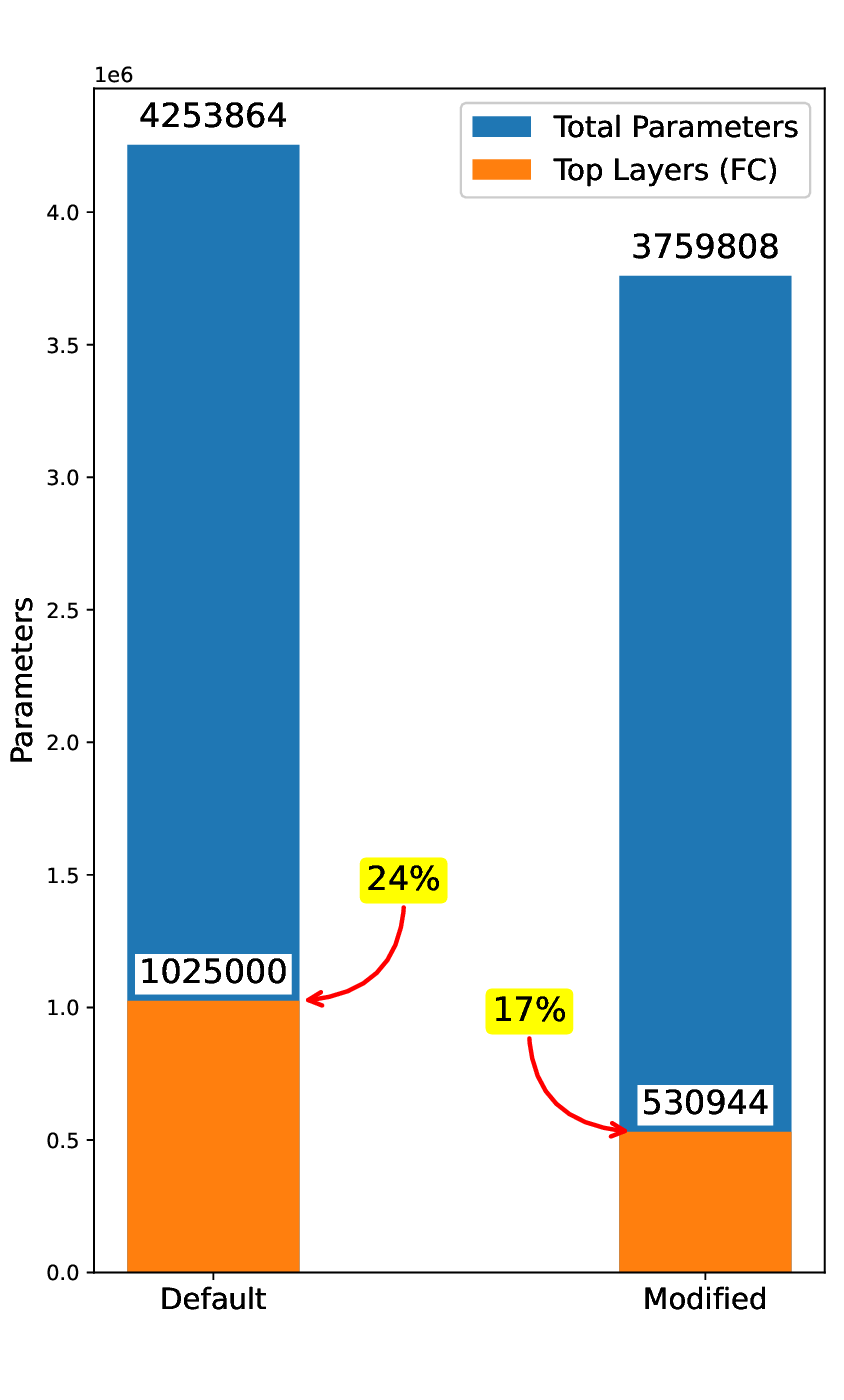} \\
		\caption{Comparison of the number of parameters in the top layer for MobileNets and M-MobileNets.}
		\label{Fig1}
	\end{figure}

The MobileNets model depends on the depth-wise separable convolutions. It is a pattern of factorized convolutions that divides a standard convolution into a depth-wise convolution and a $1\times 1$ convolution termed pointwise convolution. For MobileNets, the depth-wise convolution covers a single filter to a particular input channel. The depth-wise separable convolution divides it into two layers, a separated layer for filtering and another for combining. Acknowledging the need for reduced computation neural networks in remote environments, our M-MobileNets uses 17\% of the total parameters instead of the fully connected neural networks as depicted in Figure \ref{Fig1}, which we consider as one of the key contributions of this work.

The summary of the key contributions of this study are given below:
	\begin{enumerate}

		\item  We propose a lightweight modification of the MobileNet architecture called M-MobileNets that employs only half of the top layer parameters and achieves 97\% accuracy in fish image classification, outperforming the benchmark models.

		\item  We build a custom Indonesian fish dataset consisting of  37,462 images of 667 different species of fish that is used to train the proposed model.

		\item We demonstrate the implementation of the M-MobileNets model on low-resource hardware such as GTX 860M that is critical in remote sea environments.
  
		\item The proposed species classifier is extended to consumability classifier by pairing it with a fish database.
  
	\end{enumerate}

The remainder of this paper is organized as follows. In Section \ref{s2}, we review recent literature pertinent to the problem of deployment of deep learning technology in remote fishing areas including the issues related to remote communications, limited hardware resources, and consumability determination. 
Section \ref{s3} provides the details of the modified lightweight deep learning model M-MobileNets for fish classification. We also describe the custom fish dataset created for our study.
Section \ref{s5} presents and discusses the results of evaluation of M-MobileNets against the benchmark models. Section \ref{s6} concludes the paper with a summary of the study and recommendations for future research.

\section{Remote deployment}
\label{s2}
	
Excessive and indiscriminate fishing can lead to depletion of marine resources and extinction of species. Given the strong and growing demand for seafood, sustainable pathways for fishing and marine farming must be urgently developed. One of the key parts of the fishing industry are small fishing boats that often operate in remote locations with limited communication and computing capacity. Deploying modern technologies on these vessels poses a major challenge in implementing sustainable development strategies. In this section, we discuss some of the issues related to deployment of technology including deep learning frameworks to remote fishing areas with limited computing power.

As highlighted in the discussion below the communication technology and hardware resource challenges related to deploying software including deep learning models on fishing vessels requires new approaches. To address the issues related to remote deployment, lightweight edge intelligence models are required.

	\subsection{Communication Technology}
	The maritime communication service tends to be expensive due to the high cost of satellite transmissions and limited coverage of the terrestrial networks. It poses a challenge to approaches that rely on off-site computing including cloud-based deep learning methods. Therefore, managing and achieving efficient radio resources become critical issues in maritime communications. Huang et al. \cite{Huang2021RA} proposed a new general energy efficiency (GEE) maximization-based distributing D2D resource allocation (GEEM-DD2D-RA) scheme for maritime communication. Their scheme considered the power and interference aspects to achieve a higher energy efficiency system using less power. It is particularly beneficial for maritime out-of-coverage (OOC) D2D communications.
	
	Unmanned surface vehicles (USVs) are considered a promising technique to carry out automatic emergency tasks in the continuously changing maritime traffic environment. Nevertheless, the task allocation efficiency for USVs in the maritime environment is currently inadequate. The crucial challenge is the performance of aquatic transmission between USVs and offshore platforms. To improve the task allocation efficiency, Zhang et al. \cite{Zhang2020} proposed a state-of-the-art task allocation scheme for USVs in the smart maritime internet of things (IoT). Their results showed that the scheme has higher network resource utilization and more allocated tasks than the conventional schemes. Furthermore, they planned to establish a crowdsourcing scenario for USVs in the smart maritime to conduct sensing tasks.
	
	Similar to other industrial sectors, aquaculture substantially benefits from the deployment of Internet of Things (IoT) technologies. Adapting IoT within the aquaculture industry gives possibilities to optimize fish farming processes. Parri et al. \cite{Parri2020} proposed a real-time monitoring infrastructure using Fixed Nodes and Mobile Sinks for real-time and remotely controlling offshore sea farms. The suggested architecture takes advantage of the LoRaWAN network infrastructure for data transmission. The testing results of different configurations on the field proved that the reliability of the transmission channel in a worst-case scenario is up to an 8.33 km offshore distance. Different communication setups were evaluated to find the best compromise ratio between power consumption and data transmission reliability.
	
	One fundamental element in performing long-term endurance ocean missions in remote areas is to maintain a communication link with the systems. To deal with this issue, the BlueCom+ project presented a wireless mobile communications network with high bandwidth and tens of kilometers’ ranges by \cite{Ferreira2017}. The autonomous systems carried out data gathering and performance tests while using the BlueCom+ communications network. The tests were divided into three open sea campaigns at large Sesimbra (near Lisbon, Portugal). These trials proved that it is possible to have autonomous robotic systems with long endurance missions of patrolling/monitoring the oceans.
	
	Baharudin et al. \cite{Baharudin2012} developed a mini-grid hybrid power system to maintain a reliable clean water supply in rural areas and emergency conditions. The designed processes consist of a mini-grid power system along with a desalination plant and economic analysis on the entire project life cycle. The mini-grid power system uses solar power as the only power resource because the geographic conditions of the rural areas are not feasible for constructing transmission lines interconnected with the current National Grid. The mini-grid power system acts as a steady power supply for the desalination plant to produce clean water. The concern about economic issues is related to the initial capital cost invested, the total net present cost (NPC), the cost of electricity (COE) generated by the system per kWh, and the simple payback time (SPBT) for their project.
	
	
	\subsection{Hardware Resources}
	There are limited computing hardware resources available on remote fishing boats which impedes implementation of traditional deep learning models on board of the ships. Therefore, a more efficient model is required that can run on limited capacity processing units. Huang et al. \cite{Huang2020GPU} studied the visualization and compression of trajectories of large-scale vessels and its Graphics Processing Unit (GPU)-accelerated applications. The visualization was employed to study the effects of compression on the data quality of vessel trajectory. They applied the Douglas-Peucker (DP) and Kernel Density Estimation (KDE) algorithms for the visualization and trajectory compression that were substantially advanced through the GPU architecture's parallel computation capabilities. The study was carried out by doing a thorough experiment on the trajectory compression and visualization of large-scale AIS data which were the recording of ship migrations collected from 3 different water areas, i.e., the South Channel of Yangtze River Estuary, the Chengshan Jiao Promontory, and the Zhoushan Islands. Moreover, with the proportion of the vessel trajectories growing larger, their proposed framework will have more significance in the big data era.
	
	The Cyber-Enabled Ship (C-ES) is defined as an autonomous or remotely controlled vessel that relies on interconnected cyber-physical systems (CPS) for its operations. Those systems are inadequately protected against cyber attacks. Taking into account the critical functions provided by the systems, it is necessary to address these security challenges to ensure the ship's safety. Kavallieratos \cite{Kavallieratos2020} proposed the Maritime Architectural Framework to evaluate and portray the C-ES environment. They also applied the Secure Tropos methodology to obtain the security requirements of the vulnerable CPSs in a C-ES, which are the Automatic Identification System (AIS), the Electronic Chart Display Information System (ECDIS), and the Global Maritime Distress and Safety System (GMDSS). It was intended as a system handling the requirements of each combining system.
	
	Internet of Ships (IoS) is the network of smart interconnected maritime objects, devices, or infrastructures associated with ships, ports, or their transportation. The goal is to considerably enhance the efficiency, safety, and environmental sustainability in the shipping industry. Aslam et al. \cite{Aslam2020} provided a complete survey of the IoS paradigm, architecture, key elements, and main characteristics. Furthermore, they also reviewed the novelty of IoS applications, such as route planning and optimization, safety enhancements, decision making, automatic fault detection, cargo tracking, and preemptive maintenance, environmental monitoring, automatic berthing, and energy-efficient operations. Future challenges and opportunities for researches related to satellite communications along with its security and privacy, aquatic data collection, and management by providing a roadmap towards optimal maritime operations and autonomous shipping.
	
	The lack of infrastructures in maritime communication, i.e.,  optical fibers and base stations makes it an immensely complex and heterogeneous environment. It can also be a barrier for future service-oriented maritime IoT since it affects reliability and traffic steering efficiency. One of the promising solutions is an AI-empowered autonomous network for ocean IoT. However, AI typically involves training/learning processes and requires a realistic environment to attain beneficial outcomes. Yang et al. \cite{Yang2020} proposed the parallel network that can be viewed as the “digital twin” of the real network and responsible for realizing four key functionalities: self-learning and optimizing, state inference and network cognition, event prediction and anomaly detection, and knowledge database and snapshots. Nevertheless, critical issues remain for further study,  i.e., feature space definition, algorithm selection, and evaluation, and coping with errors.

	\subsection{Consumability Determination}
	Edibility is one of the primary factors in fish categorization. Beyond classifying the species of fish, it is also important to determine whether the captured fish is consumable. Thus, fish evaluation should consist of two parts: i) species classification, and ii) consumability indicator. A fish classifying system has been developed by Winiarti et al. \cite{winiarti2020} by using the k-Nearest Neighbor (kNN) as the classifier to segregate consumable fishes into four classes based on its texture extraction color features. The fish's meat and scales are used as identification parameters. The fish meat is captured by the HSV colors model (hue, saturation, and value) and GLCM (Gray Level Co-occurrence Matrix) method, and the values are used for the scales' texture feature extraction.  The accuracy for the scales reached 87.5\% for tilapia and 95\% for mackerel.
	
	The performance of various FC techniques relies on the pre-processing and feature extraction methods, the amount of extracted features and the accuracy of the classification, the counts of fish families/species recognized. Reference \cite{Alsmadi2020} evaluated database usages such as Fish4-Knowledge (F4K), knowledge database, and Global Information System (GIS) on Fishes and others. They also studied the preprocessing methods features, extraction techniques, and classifiers from previous works to understand its characteristics as guidance for future research and fulfill the current research gaps. Their study concluded that the most commonly used algorithms for classification were VM, BP algorithm, HGAGD-BPC, GAILS-BPC, Bayesian classifier, and CNN.
	
	Environmental variations such as luminosity, fish camouflage, dynamic backgrounds, water murkiness, low resolution, the swimming fish's shape deformations, and subtle differences between fish species give challenges in underwater videos. Jalal et al. \cite{Jalal2020} proposed a hybrid solution to overcome these challenges by combining optical flow and Gaussian mixture models with YOLO deep neural network as an approach to unconstrained underwater videos for detecting and classifying fishes. YOLO-based object detection systems are originally capable of capturing only the statistic and visible fish details. They eliminated this limitation to enable YOLO to detect moving fish or camouflaged fish, by utilizing temporal information obtained from Gaussian mixture models and optical flow. The suggested system was evaluated on underwater video datasets i.e., the LifeCLEF 2015 from the Fish4Knowledge and a dataset from The University of Western Australia (UWA).
	
	In most fisheries, the length of fish is still measured manually. The results give precise length estimation at fish level but the sample size tends to be small because of the high inherent costs of manual sampling. Alvarez-Ellacuria et al. \cite{Alvarez2019} presented another approach for fish measurement by using a deep convolutional network (Mask R-CNN) for automatic European hake length estimation from automatically collected images of fish boxes. The results give average lengths ranging from 20–40 cm, the root-mean-square deviation was 1.9 cm, and the maximum deviation between the estimated and the measured mean body length was 4.0 cm. The estimated mean of fish lengths is accurate at the box level, however, the species detection from the same image is still needed to be addressed.
	
	River systems are formed by disruptions of floods and droughts, hence, the river fish species have evolved features to make them more resilient to the disruption. Treeck et al. \cite{VanTreeck2020} have analyzed and summarized the resilience features of European lampreys and fish species to acquire a unique species sensitivity classification to mortality. They have gathered the fish's features such as maximum length, migration type, mortality, fecundity, age at maturity, and generation time of 168 species and developed an original method to weigh and integrate those features to create each species' final sensitivity score ranging from one (low sensitivity) to three (high sensitivity). Large-bodied, diadromous, rheophilic, and lithophilic species such as Atlantic salmons, sturgeons, and sea trouts usually have a higher sensitivity to additional adult fish mortality than the small-bodied, limnophilic, and phytophilic species with fast generation cycles. The final score and classification can be easily localized by picking the most sensitive species to the local species pool.

	\section{Fish Classification Model}
	\label{s3}
	In this section, we discuss the main components of the proposed approach for efficient fish classification including i) data collection, ii) data augmentation, iii) the modification of the MobileNet architecture to obtain a more lightweight deep learning model, and iv) transfer learning.

	\subsection{Data}

 One of the key parts of our study is the collection of a custom image dataset of Indonesian fish. A network of fishing boats operating in the waters of the Indonesian archipelago was contracted to help obtain and label the images of fish. Fishing crews were instructed to take photos of captured fish together with its measurements. Each specimen was also assigned its species label by the fishermen. Thus, a total of 37,462 images of fish was collected.

 	The original images taken on board of the fishing boats consisted of different dimensions due to different photo equipment used by the crew. During the processing stage the images were homogenized and scaled to \texttt{224$\times$224} (pixel). The original images consisted of the RGB values in the range \texttt{0-255} which was adjusted to the range \texttt{0-1}  through rescaling. Otherwise, the RGB values would be too high for the model to process in low-resource computation. As some level of diversity could make the data suitable for the upcoming unseen data, the images of the fishes were taken randomly. Some images were taken under the water while others were taken outside the water with various unspecified angles and distances.
  
 The images are categorized based on their species, genus, family, and order as can be seen in Table \ref{tab:1}. The proposed model is trained on the final dataset to classify the species of fish. The data is split into training and test sets containing 29,970 and 7,492 images, respectively. The images were categorized into 283 genera of fish from 667 species. Furthermore, the data was synced with FishBase \cite{FishBase} - a provider of fish information around the world - to determine whether each specimen is consumable or not.
	
	\begin{table*}[ht]
		\tiny
		\caption{Preview of the full fish dataset containing 667 species.}
		\label{tab:1}
		\hskip-1.5cm
		\begin{tabular}{ | l | l | l | l | l | l | l | l | c | c | } \hline
			\rowcolor{SkyBlue}\textbf{ID} & \textbf{Order} & \textbf{Family} & \textbf{Genus} & \textbf{Species} & \textbf{Occurrence} & \textbf{Foreign name} & \textbf{Local Name} & \textbf{Description} & \textbf{Category} \\ \hline
			0 & Tetraodontiformes & Balistidae & Abalistes & Abalistes stellaris & native  & Starry triggerfish  &   & commercial  & Commercial \\ \hline
			$\ldots$  & $\ldots$ & $\ldots$ & $\ldots$ & $\ldots$ & $\ldots$  & $\ldots$  & $\ldots$  & $\ldots$  & $\ldots$ \\ \hline
			7 & Perciformes & Acanthuridae & Acanthurus & Acanthurus leucosternon & native  & Powderblue surgeonfish  & Botana biru  & minor commercial  & Commercial \\ \hline
			13 & Perciformes &  &  & Acanthurus olivaceus & native  & Orangespot surgeonfish  & Botana coklat  & commercial  & Commercial \\ \hline
			14 & Perciformes &  &  & Acanthurus pyroferus & native  & Chocolate surgeonfish  & Botana model  & commercial  & Commercial \\ \hline
			17 & Perciformes & Serranidae & Aethaloperca & Aethaloperca rogaa & native  & Redmouth grouper  & Geurape itam  & minor commercial  & Commercial \\ \hline
			27 & Perciformes & Carangidae & Alepes & Alepes vari & native  & Herring scad  & Trevally scad  & minor commercial  & Commercial \\ \hline
			34 & Perciformes & Pomacentridae & Amphiprion & Amphiprion clarkii & native  & Yellowtail clownfish  & Giro pasir biasa  & subsistence fisheries  & Commercial \\ \hline
			35 & Perciformes &  &  & Amphiprion melanopus & native  & Fire clownfish  & Black anemonefish  & subsistence fisheries  & Commercial \\ \hline
			36 & Tetraodontiformes & Tetraodontidae & Arothron & Arothron hispidus & native  & White-spotted puffer  & Stars-and-stripes pufferfish  & poisonous to eat  & Danger \\ \hline
			44 & Tetraodontiformes &  &  & Arothron immaculatus & native  & Immaculate puffer  & Buntel pasir  & poisonous to eat  & Danger \\ \hline
			59 & Tetraodontiformes &  &  & Arothron mappa & native  & Map puffer  & Scribbled pufferfish  & poisonous to eat  & Danger \\ \hline
			61 & Tetraodontiformes &  &  & Arothron meleagris & native  & Guineafowl puffer  & Guinea-fowl pufferfish  & poisonous to eat  & Danger \\ \hline
			74 & Tetraodontiformes &  &  & Arothron nigropunctatus & native  & Blackspotted puffer  & Buntel babi  & poisonous to eat  & Danger \\ \hline
			75 & Tetraodontiformes &  &  & Arothron stellatus & native  & Stellate puffer  & Starry pufferfish  & poisonous to eat  & Danger \\ \hline
			76 & Tetraodontiformes & Balistidae & Balistapus & Balistapus undulatus & native  & Orange-lined triggerfish  & Triger lorek  & traumatogenic  & Danger \\ \hline
			77 & Tetraodontiformes &  & Balistoides & Balistoides viridescens & native  & Titan triggerfish  & Lubien manok  & reports of ciguatera poisoning  & Danger \\ \hline
			$\ldots$  & $\ldots$ & $\ldots$ & $\ldots$ & $\ldots$ & $\ldots$  & $\ldots$  & $\ldots$  & $\ldots$  & $\ldots$ \\ \hline
		\end{tabular}
	\end{table*}

	\subsection{Data Augmentation}
	To improve classifier performance on unseen data we augment the original training set with various additional modified images. The goal of data augmentation is to expose the classifier to a larger variety of images which should improve classifier robustness to various image distortions. The augmentation techniques might be imperceptible for humans, however, it gives significant alteration for the computing device to process because the arrays are also changing. The data was augmented using following transformations:
	
	\begin{itemize}
		\item Rescale, the original image has RGB coefficients that make low-resource devices unable to process, so the image is rescaled to get RGB coefficient between 0 and 1.
		\item Width Shifting, shifting images to the left or right with a pre-determined range of floating-point numbers between 0.0 and 1.0. The range specifies the fraction of boundary level of the total width.
		\item Height Shifting, works exactly like width shifting but in vertical.
		\item Shear, a different rotation technique to transform images by stretching them on a certain angle known as the shear angle.
		\item Zoom, a magnifying image at a specific range of floating-point numbers. A number less than 1.0 magnifies the image, while a number greater than 1.0 zooms out the image
		\item Flip, this technique will flip the image either horizontally or vertically
		\item Fill, this technique will repeat the closest pixel of a certain pixel and fill the empty values.
	\end{itemize}

	\subsection{Mechanism}

\subsubsection{MobileNets}
MobileNets was originally developed by Google. It is based on a streamlined architecture that uses depth-wise separable convolutions to build lightweight deep neural networks. It was introduced as an efficient deep learning model for mobile and embedded vision applications. MobileNets efficiency and low resources usage led to its adoption on mobile devices like smartphones. Given the limited computing capacity on board of remote fishing boats, MobileNet provides an attractive base model to implement our fish classification model.

\subsubsection{Modified MobileNets}
\begin{figure*}[!ht]
		\centering
		\includegraphics [width=0.99\textwidth]{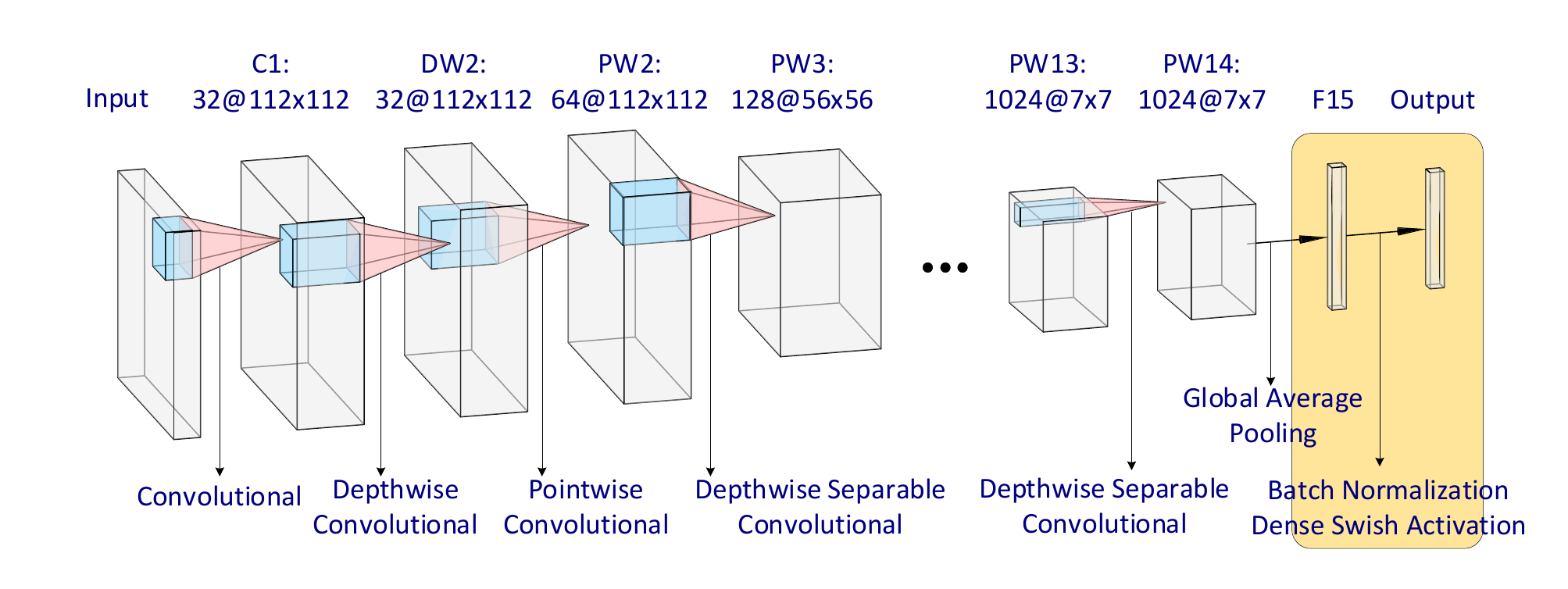} \\
		\caption {The architecture design of M-MobileNets.}
		\label{Fig5}
	\end{figure*}

We modify the original MobileNet architecture to fit our purposes in the context of resource constrained optimization. To this end, several changes to the original model are implemented including reducing the number of top layer parameters (Figure \ref{Fig1}), using a new activation function (Swish), and introducing batch normalization within the model. The new modified model is called M-MobileNets. The final architecture of the proposed classification model is shown in Figure \ref{Fig5}.

To construct our proposed modified MobileNet model (M-MobileNet), we reduce the number of parameters in the top layer while keeping the CNN layers unchanged. Concretely, the CNN layers of M-MobileNets are kept exactly the same as the original MobileNets, while the number of parameters in the fully connected top layers of M-MobileNets is reduced to around 531,000 compared to around 1,025,000 in the original MobileNets model. Thus, we obtain a lighter version of the original model that is faster and smaller. Due to its reduced size, M-MobileNets can be employed on edge computing devices.

The second key modification is the use of a new Swish activation function $S(x)$ which was originally introduced by Google. It has been known to surpass ReLu in various scenarios. The choice of the activation function used in the network affects the training dynamics and can improve classification performance. The Swish activation is closely related to the traditional sigmoid activation $\sigma(x)$ function.
	
	\begin{equation}
		\label{eq:1}
		S(x):= x \times \sigma (\beta x) = \frac{x}{1+e^{-\beta x}},
	\end{equation}

	The utility of the Swish activation can be optimized when used in conjunction with batch normalization which has gradient squishing property. Batch normalization allows faster and more stable training of the neural net through normalization of the layers' inputs by re-centering and re-scaling
	
	Normalization is performed when $S(x)$ goes through a mini-batch of size $m$ with mean $\mu_B = \frac{1}{m} \sum_{i=1}^{m}x_i$ and variance $\sigma_B^2 = \frac{1}{m} \sum_{i=1}^{m}(x_i - \mu_B)^2$. The inputs of each layers will be separately normalized and denoted by
	$\hat{x}_i^{(k)} = \frac{\hat{x}_i^{(k)}-\hat{x}_B^{(k)}}{\sqrt{\sigma_B^{(k)^2}+\epsilon}}$ where $k \in [1,d]$ and $i \in [1,m]$ (d is dimension and m is mini-batch) respectively while $\mu_B^{(k)}$ and $\sigma_B^{(k)^2}$ are mean and variance for each dimension.

\subsubsection{Categorization}
The proposed method categorizes fish in two ways. First, the M-MobileNet model classifies the species of the input image. In other words, given an image of a fish the model determines its species. In the second stage, the species information is fed to FishBase to determine if the fish is edible. 
The classification is intended for various types of fishes with binary labels of consumable and unconsumable (dangerous). Consumable fish dominate the commercial fishes and most unconsumable fish are dangerous fishes that are poisonous. To gather information regarding fish species, the dataset refers to the FishBase where fish characteristics have been mapped according to their species, genus, family and order. Based on the conducted experiments, the classification could be more efficient if the fishes are mapped by their genera rather than species because many species have similar characteristics but with different labels. For example, the Carcharhinus Limbatus and Carcharhinus Melanopterus, two types of sharks indistinguishable by their physical characteristics. It could confuse the model to classify them in species class. But, since both of them are sharks and are part of the unconsumable fish category, the model should classify them in their genera (Carcharhinus) as shark types.
	
	These fish labels are hard to determine by appearance only, thus, the classification must be undertaken through species information first and then goes to the labels. The proposed mechanism of fish classification to determine its consumable features aims for efficiency in low resource/smaller devices (i.e Raspberry pi, smartphone).
	
\begin{figure}[!ht]
		\centering
		\includegraphics [width=0.48\textwidth]{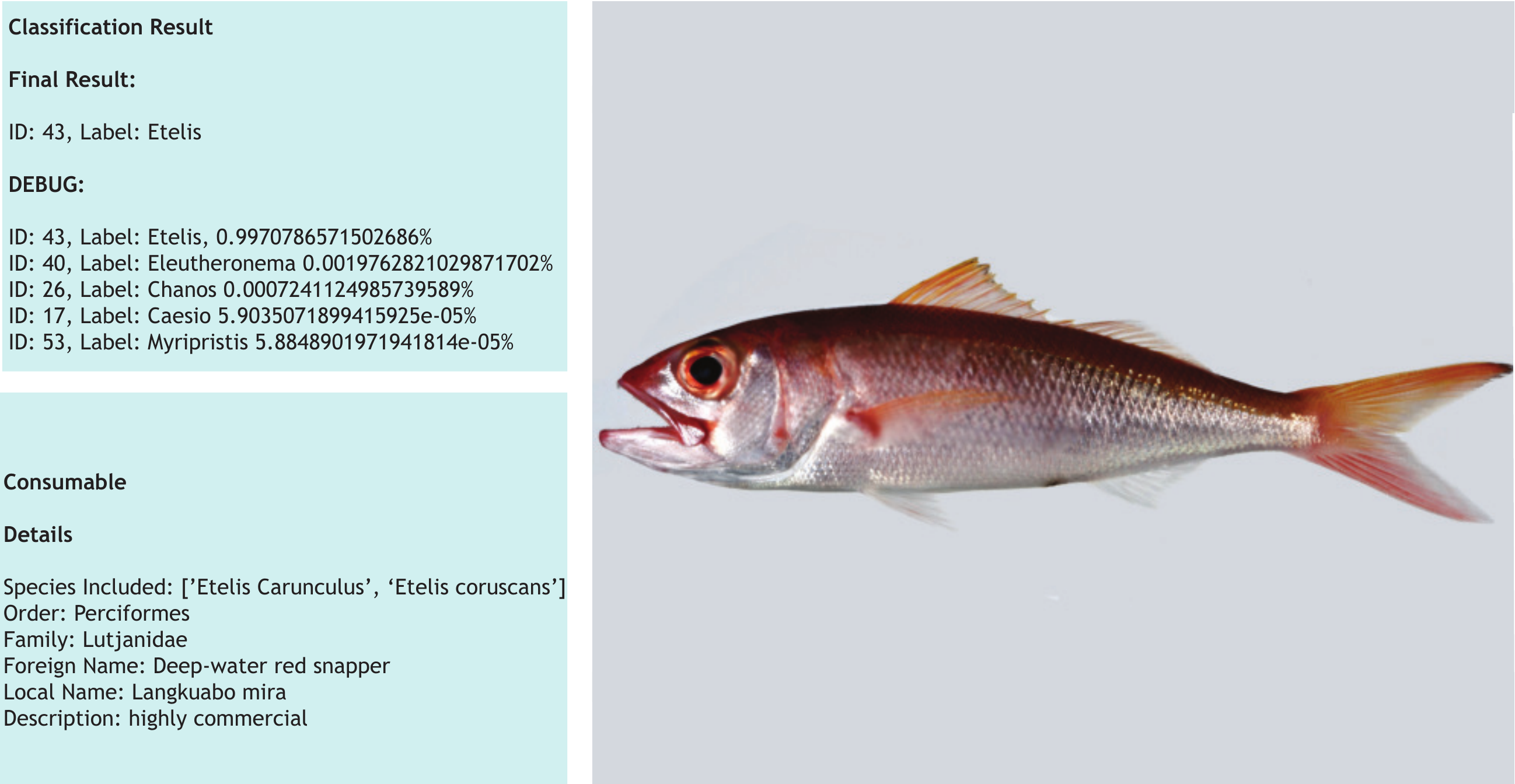} \\
		\caption {Example of final classification result: \textit{Consumable}}
		\label{Fig3}
	\end{figure}

\begin{figure}[!ht]
		\centering
		\includegraphics [width=0.48\textwidth]{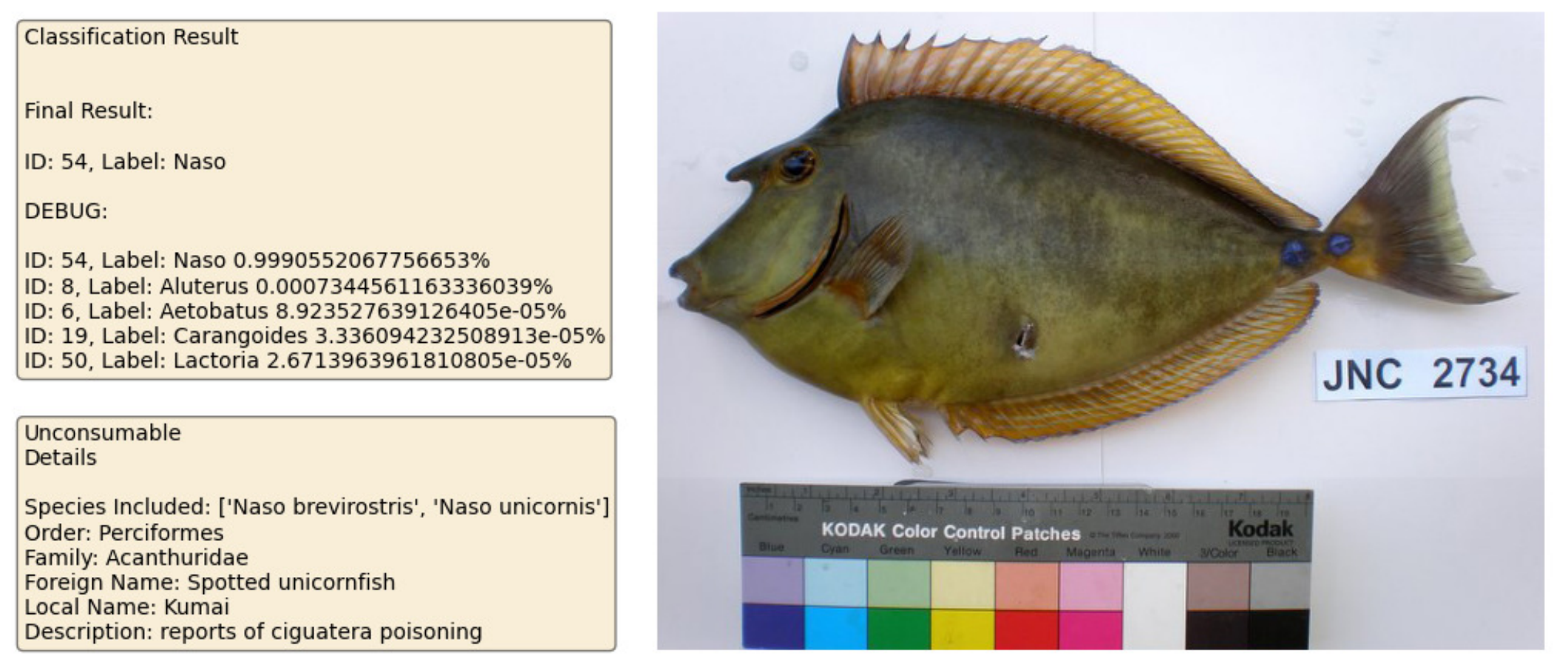} \\
		\caption {Example of final classification result: \textit{Unconsumable}}
		\label{Fig4}
	\end{figure}

	A fish image is fed into the model classification which gathers information related to the fish based on its appearances. The output is then forwarded to the label classification that is connected to the fish database to determine whether the fish fits in \textit{Consumable} category (as can be seen in Figure \ref{Fig3}) or \textit{Unconsumable} category (as can be seen in Figure \ref{Fig4}).

\subsection{Transfer Learning}
\label{s4}
Transfer learning is a machine learning technique for recycling an existing trained model for use in another task. This approach provides optimization and rapid progress on modeling the second task. It is a good way to save resources especially on a problem in which the input is image type data. The proposed method and benchmark results use the well-known weight from Keras which is developed with the ImageNet dataset. The general features learned from the ImageNet dataset indeed help the development of the fish classifier model, but a transfer learning model for a second task still needs to be tuned for optimized performance and accuracy. The trial and error are conducted to find the most adaptive and optimal configuration for the model, which results in the choice  of the Swish activation. Transfer learning suits the low-resource scenario because of the efficiency in time and accuracy of the fish dataset. 

Another important parameter that requires additional consideration and tuning during transfer learning is the  optimal learning rate. The study eventually used the $10^{-4}$ value of the learning rate. Determining the optimal learning rate holds the key to the model's accuracy while making it faster. The trial started with a large value i.e., 0.1 then it lowered exponentially. A large learning rate might cause the model to train faster however it will not be able to reach the optimal accuracy. Meanwhile, a smaller learning rate might slow down the model training.

\section{Evaluation and Analysis}
\label{s5}

To evaluate the proposed approach for fish classification, we benchmark it against several existing models. Model evaluation is done based on the metrics  precision, recall, sensitivity, F-score for both micro averages and macro averages along with  accuracy. To measure the hardware performance, we compare the GPU utility on the proposed and benchmark models.

	\subsection{Performance Evaluation}
 The performance of each model is measured based on the confusion matrix which is derived from the comparison of the true and predicted labels. The confusion matrix is used calculate the key performance metrics including precision, recall (sensitivity), specificity, F-score, and accuracy which is represented by the following equations \cite{powers2020,ZHOU2019}:
	
	\begin{equation}
		\label{eq:6}
		Precision = \frac{TP}{TP + FP}
	\end{equation}
	
	\begin{equation}
		\label{eq:7}
		Recall = \frac{TP}{TP + FN} = \frac{TN}{P}
	\end{equation}
	
	\begin{equation}
		\label{eq:8}
		Specificity = \frac{TN}{TN + FP} = \frac{TN}{N}
	\end{equation}
	
	\begin{equation}
		\label{eq:9}
		F_\beta = \frac{(1+\beta^2)(PREC\cdot REC)}{(\beta^2 \cdot PREC+REC)}
	\end{equation}
	
	\begin{equation}
		\label{eq:10}
		Accuracy = \frac{TP + TN}{TP + FP + TN + FN}
	\end{equation}

	\subsection{Activation analysis}
\begin{figure}[!t]
		\centering
		\includegraphics [width=0.45\textwidth]{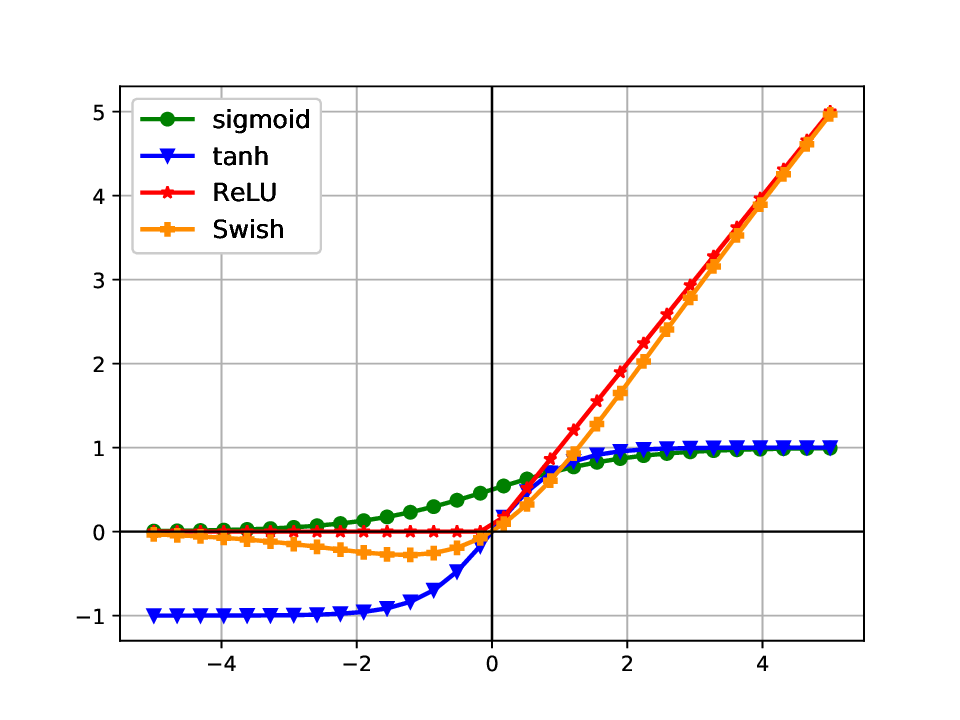} \\
		\caption {Activation functions considered in the study.}
		\label{Fig6}
	\end{figure}

One of the key components in neural network architecture is the activation function. The activation function plays an important role in transmitting the gradient signal through the network during the learning stage. A poor activation function can hinder effective learning even if all other components of the pipeline are in place.
We compare the performance of different activation functions to determine the optimal activation. In particular, we consider sigmoid ($\sigma$), $\tanh$, ReLU, and Swish activations (Figure \ref{Fig6}). 

The sigmoid function is popular for its smooth probabilistic shape with an equation:
	
	\begin{equation}
		\label{eq:2}
		\sigma(x) = \frac{1}{1+e^{-x}}
	\end{equation}
	
	It is a convenient way to efficiently calculate gradients in a neural network. On the other hand, sigmoid function flattens rather quickly. The values go to $0$ or $1$ instantly causing the partial derivatives to quickly go to zero and the resulting weights cannot be updated which makes the model unable to learn. The $\tanh$ activation can be viewed as a scaled version of the sigmoid and has similar gradient issues as the  sigmoid function. The equation is given below:
	
	\begin{equation}
		\label{eq:3}
		\tanh(x) = \frac{e^x-e^{-x}}{e^x+e^{-x}} = 2 \cdot \sigma(2x)-1
	\end{equation}
 
 As can be seen in Figure \ref{Fig6}, the Swish activation is unbounded on the positive $x$-axis. It means that for high values for $x$,  the learning process is improved over $\sigma$ and $\tanh$. Moreover, Swish is non-monotonic, which means that Swish has both negative and positive derivatives at some point. This increases the information storage capacity and the discriminative capacity of the model. 
	
One of the most popular activation functions is Rectified Linear Unit (ReLU) which is given by the following equation:
	
	\begin{equation}
		\label{eq:4}
		f(x) = \begin{cases}
			0 & \text{if  $x <0$}\\
			x & \text{if $x \geq 0$}
		\end{cases}
	\end{equation}

\begin{figure}[!ht]
		\centering
		\includegraphics [width=0.48\textwidth]{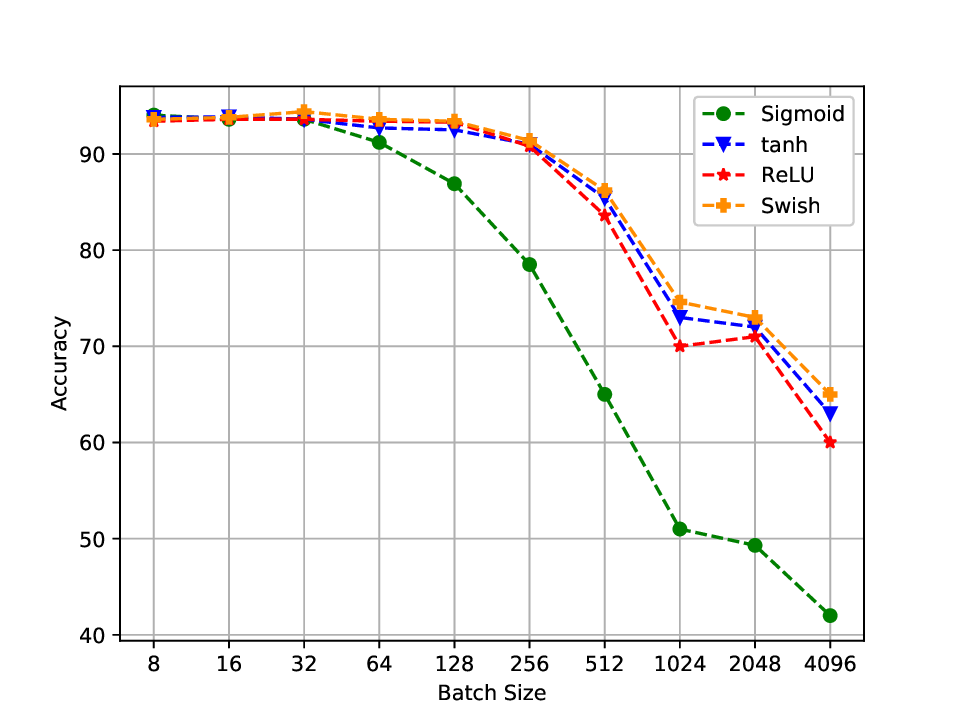} \\
		\caption {Accuracy comparison of activation functions trained with 50 epochs.}
		\label{Fig9}
	\end{figure}

While the ReLU activation resembles Swish activation for positive values of $x$ (Figure \ref{Fig6}), it exhibits different behavior for the negative values. The non-monotonic nature of Swish sets it apart from ReLU and other activations. As shown in Figure \ref{Fig9}, Swish outperforms ReLU across different batch sizes. 
	
\begin{figure}[!t]
		\centering
		\includegraphics [width=0.45\textwidth]{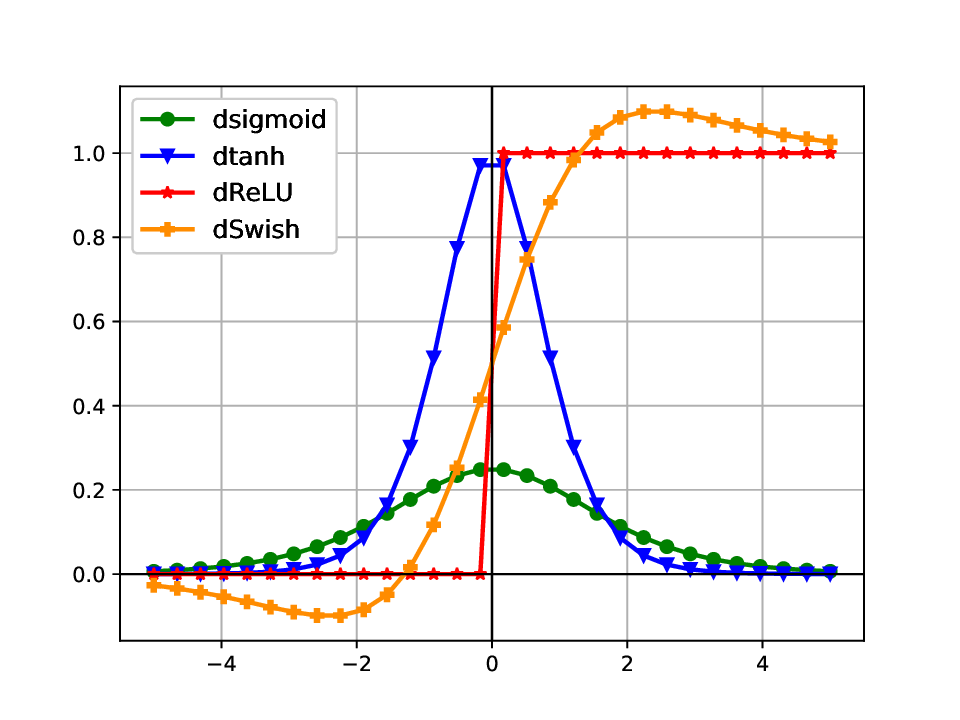} \\
		\caption {Derivative of the activation functions considered in the study.}
		\label{Fig7}
	\end{figure}

To better understand the behavior of the activation functions during the training phase of a neural network, it is instructive to consider the derivative of the functions which are depicted in Figure \ref{Fig7}.

In particular, the derivative of Swish activation is symmetric and reduces the phenomenon of vanishing gradient. The derivative has an interesting property given by the following equation:
\begin{equation}
    S'(x)=S(x)+\sigma(x)(1-S(x))
\end{equation}

The results of the comparison of activation functions are presented in Figure \ref{Fig9}. It shows that Swish activation produces the highest accuracy among all the tested activations across all the batch sizes. Therefore, we choose Swish as the main activation in the proposed modified MobileNets model.

	\subsection{Main Results}
 In this section, we present and discuss the main results of the comparison between the proposed M-MobileNets and the benchmark methods. The models are compared based on several classification metrics as well as the GPU performance. Given the context in which the proposed model is to be deployed both classification accuracy and low computational overhead are important factors in evaluating the models.
 
 The results of the classification metrics are shown in Table \ref{tab:2}. It can be seen that M-MobileNets outperforms the benchmarks across all the criteria - precision, recall, F1-score, and specificty. Most importantly, M-MobileNets achieves the highest accuracy of 97\% which is significantly better than the benchmarks. M-MobileNets achieves the best results both in terms of micro and macro averages. The dominant results support the superiority of the proposed model.

	\begin{table*}[ht]
		\centering
		\footnotesize
		\caption{Classification metrics for M-MobileNets and the benchmark models.}
		\label{tab:2}
		\begin{tabular}{ | l | c | c | c | c | c | c | c | c | c | } \hline
			& \multicolumn{3}{c}\textbf{Micro Averages} &  & \multicolumn{3}{c}\textbf{Macro Averages} &  &  \  \\ \hline
			\textbf{Architecture} & \textbf{Precision} & \textbf{Recall} & \textbf{F1-Score} & \textbf{Specificity} & \textbf{Precision} & \textbf{Recall} & \textbf{F1-Score} & \textbf{Specificity} & \textbf{Accuracy} \\ \hline
			VGG16 		& 0.949	& 0.949 &	0.949 &	0.999	& 0.944	& 0.923	& 0.933 &	0.999	& 95  	\\ \hline
			Resnet50 	& 0.932	& 0.931 &	0.931 &	0.999	& 0.929	& 0.928	& 0.929 &	0.999	& 93	\\ \hline
			\rowcolor{GreenYellow}M-MobileNets & 0.978	& 0.978 &	0.978 &	0.999		& 0.942	& 0.960	& 0.951 &	0.999	& 97  	\\ \hline
			MobileNetv2 & 0.936	& 0.936 &	0.936 &	0.999	& 0.885	& 0.910	& 0.999 &	0.897	& 93	\\ \hline
			\rowcolor{Apricot}MobileNets 	& 0.948	& 0.948 &	0.948 &	0.999	& 0.939	& 0.945		& 0.942 &	0.999		& 94    	\\ \hline
			Effnet  	& 0.947	& 0.947 &	0.947 &	0.999		& 0.912	& 0.919	& 0.915 &	0.999	& 94  	\\ \hline
			Capsnet 	& 0.822	& 0.822 &	0.822 &	0.998	& 0.802		& 0.787	& 0.794 &	0.998	& 82  	\\ \hline
		\end{tabular}
	\end{table*}

 Since the proposed model is designed for application on remote fishing vessels, the GPU performance plays an important role in determining the feasibility of the proposed approach. To this end, we compare the GPU utility, memory utility, and memory usage between M-MobileNets and the benchmark methods.  The GPU performance was measured by using a mobile GPU (Nvidia GTX 860M) with a considerably low memory of 4GB GDDR5. The augmentation methods were used for determining whether the condition affects GPU performance or not in each architecture. The benchmarking processes were conducted with \texttt{nvidia-smi} as the main system management interface for NVIDIA in Linux operating system.
 The results are presented in \ref{tab:3}. The results show that M-MobileNets performs well across all three criteria. In particular, it achieves the minimum memory utility and near-minimum GPU utility. 
	
	\begin{table}[ht]
		\centering
		\scriptsize
		\caption{Comparison of GPU metrics for M-MobileNets and the benchmark models.}
		\label{tab:3}
		\begin{tabular}{ | l | p{1.58cm} | p{1.65cm} | p{1.71cm} |} \hline
			\textbf{Architecture}	& \textbf{GPU Utility (Average in \%)} & \textbf{Memory Utility (Average in \%)} & \textbf{Memory Usage ( Average in MB)}  \\ \hline
			VGG16 			& 97.999	& 65.414	& 3995.168  \\ \hline
			Resnet50 		& 84.529	& 56.154	& 3996.651	\\ \hline
			\rowcolor{GreenYellow}M-MobileNets		& 42.959	& 14.567	& 4020      \\ \hline
			MobileNetv2 	& 42.434	& 15.879	& 3991    	\\ \hline
			\rowcolor{Apricot}MobileNets 		& 43.675	& 14.986	& 3816.324  \\ \hline
			Effnet  		& 52.076	& 23.253	& 4007.326  \\ \hline
			Capsnet 		& 98.982	& 68.238	& 3988.982  \\ \hline
		\end{tabular}
	\end{table}

\section{Concluding Remarks}
\label{s6}
This research was conducted to provide an efficient methodology that could help increase productivity in the marine field by utilizing a deep learning-based approach with low resources. The collected images consisted of 667 species from 37,462 images with various dimensions that were successfully loaded onto the model, with help of Kesatuan Nelayan Tradisional Indonesia (KNTI). M-MobileNets were used as the mobile architecture with a modification of the full-connected layer to get lower computation and higher accuracy. Also, the pre-built weight of ImageNet used on the model drove the training process to use generic features inside the weight to recognize the fish. It made the training go faster with fewer resources in gaining optimal accuracy. We presented M-MobileNets architecture and compared it with various architectures (e.g. mobile and non-mobile). The results showed that our proposed architecture outperforms other architectures both in micro averages and macro averages. Furthermore, with a low GPU specification of GTX 860m, the M-MobileNets usage decreased the clock usage while at the same time increasing the memory usage for gaining more accuracy compared to the conventional MobileNets architecture.
	
The assessment will be more eligible by complementing the dataset and model with a high specification camera capable of capturing multiple frames per second through a resource-constrained IoT device. The aim is to create a \textit{Single Shot Detector (SSD)} for the field implementation of the model and to maintain its performance in low-resource hardware by gaining smoothness in object detection. This can be achieved by deploying the model to continuously classify captured frames despite the limited computation power. 

\bibliographystyle{plain}
\bibliography{references.bib}

\end{document}